\documentclass{article}
\usepackage{spconf,amsmath,graphicx,multirow}
\usepackage[table,xcdraw]{xcolor}
\usepackage{pifont}

\usepackage{array}
\usepackage{amssymb}
\usepackage[colorlinks=false]{hyperref}

\title{Adaptive Graph Convolution Module for Salient Object Detection}
\name{Yongwoo Lee$^1$\quad Minhyeok Lee$^1$\quad Suhwan Cho$^1$\quad Sangyoun Lee$^{1,2}$}
\address{$^1$~Yonsei University\\$^2$~Korea Institute of Science and Technology (KIST)}

\begin{document}
\maketitle
\begin{abstract}
Salient object detection (SOD) is a task that involves identifying and segmenting the most visually prominent object in an image. Existing solutions can accomplish this use a multi-scale feature fusion mechanism to detect the global context of an image. However, as there is no consideration of the structures in the image nor the relations between distant pixels, conventional methods cannot deal with complex scenes effectively. In this paper, we propose an adaptive graph convolution module (AGCM) to overcome these limitations. Prototype features are initially extracted from the input image using a learnable region generation layer that spatially groups features in the image. The prototype features are then refined by propagating information between them based on a graph architecture, where each feature is regarded as a node. Experimental results show that the proposed AGCM dramatically improves the SOD performance both quantitatively and quantitatively. 
\end{abstract}

\section{Introduction}
Salient object detection (SOD) identifies and segments an image's most visually prominent object. The primary purpose of SOD is to automatically highlight the regions of an image that are most noticeable to a human observer. SOD is a vital pre-processing step for various computer vision tasks such as object detection and segmentation.

Because the goal is to find the most salient object of a scene, extracting global context from the input image is a crucial factor in SOD. Existing methods ~\cite{FPN,wang2019salient,park2022saliency,qin2020u2} focus on leveraging multi-scale features to obtain comprehensive representations of an image. For instance, FPN~\cite{FPN} used a pyramid structure with bottom-up and top-down pathways, and lateral connections to extract multi-scale features. Similarly, Page-Net~\cite{wang2019salient} proposed an attentive pyramid module to increase the receptive field size and used multi-scale features by repeatedly downsampling the feature maps. Park et al.~\cite{park2022saliency} introduces a context fusion decoder network consisting of a context module to extract global information of an image and a feature fusion module to refine the features from encoder and decoder. U2-Net~\cite{qin2020u2} employs a residual U-block consisting of a nested U-structure to extract multi-scale properties and enable the network to extract the feature map without degrading the feature map resolution.

\begin{figure}[t]
    \centering
    \includegraphics[width=1\linewidth]{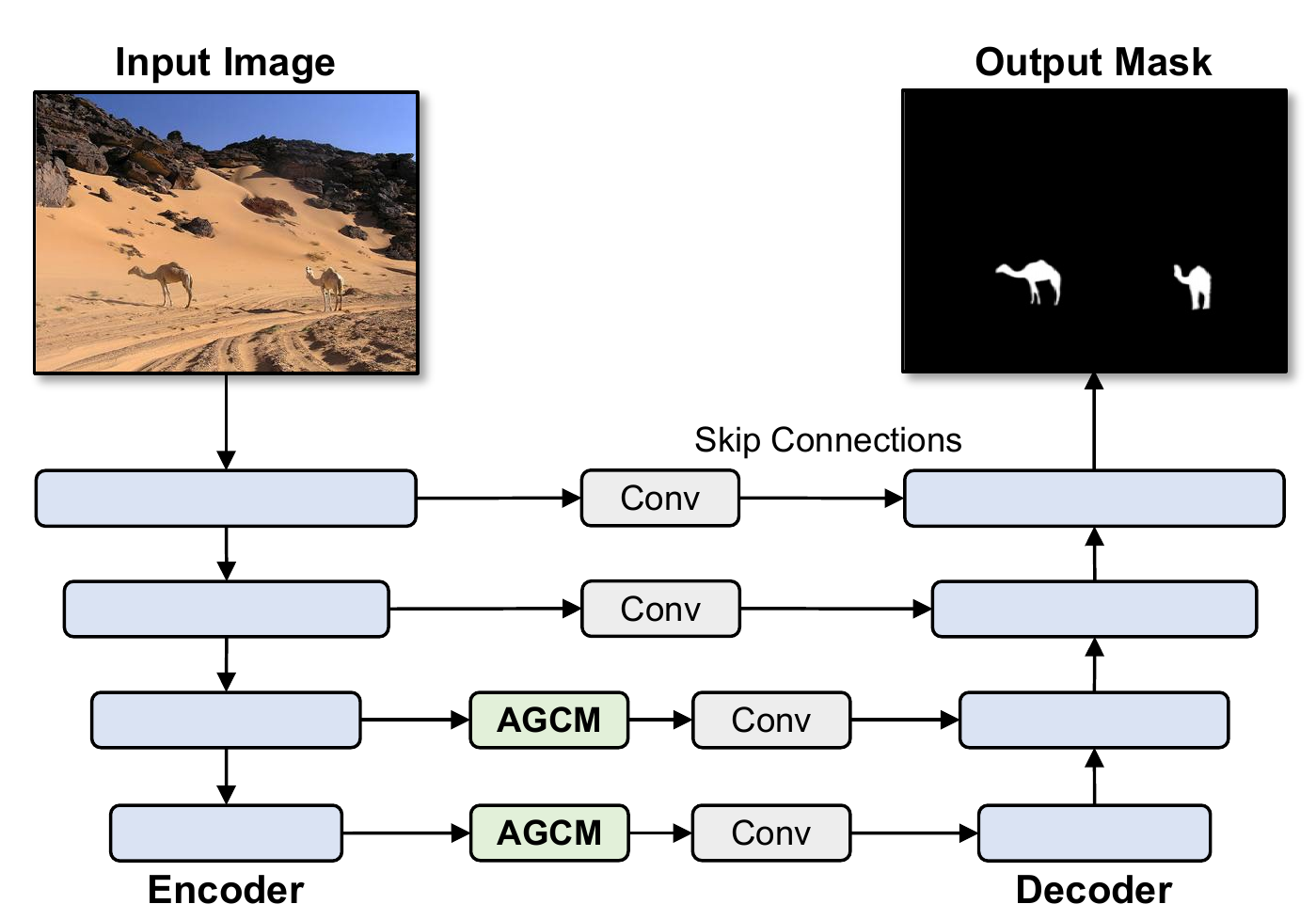}
    \caption{Visualized architecture of our model. Based on an encoder--decoder architecture, AGCM is added in the middle of skip connections, allowing the network to capture the long-range dependencies between the pixels.}
    \label{figure1}
\end{figure}

\begin{figure*}[t]
    \centering
    \includegraphics[width=0.85\linewidth]{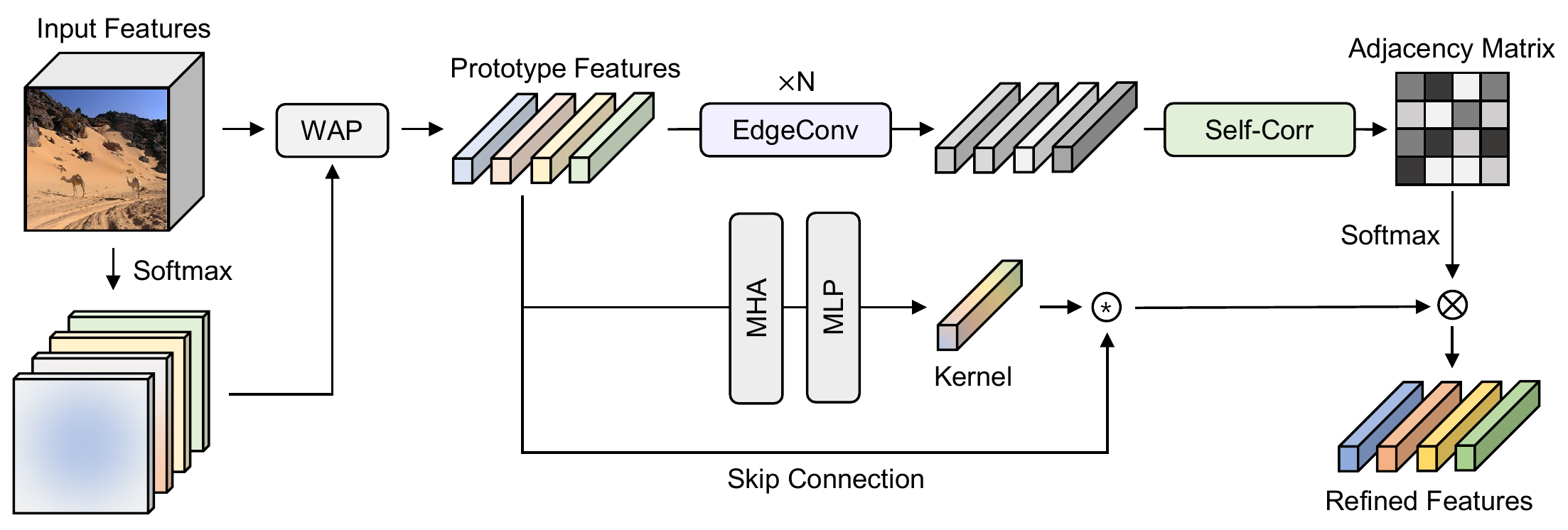}
    \caption{Pipeline of our proposed AGCM. Prototype features are first extracted from input features using a masked average pooling layer. Then, based on a graph architecture, prototype features are refined by considering each feature vector as a node. The refined features are used to calculate correlation scores with input features.}
    \label{figure2}
\end{figure*}

Although leveraging multi-scale features may help capture global context, the previous methods cannot effectively deal with complex scenes where solid understanding of the structure is required. To overcome this limitation, we propose an adaptive graph convolution module (AGCM). Inspired by a previously published graph convolution network~\cite{gnn}, we designed AGCM to preserve permutation invariance even if the order of the nodes is altered. The proposed AGCM can be divided into two stages: 1) prototype feature generation; 2) prototype feature refinement. First, given the input features, spatial regions are first generated to construct prototype feature vectors, each containing global feature representations for respective regions. In other words, the input features are semantically and spatially grouped to provide explicit structural prior of an image. This process is completely differentiable, and therefore, it can be learned in an end-to-end manner. Second, AGCM adaptively creates a weight kernel from the generated prototypes and forms an adjacency matrix between them. In particular, when forming an adaptive graph, the constructed prototype feature vectors are refined using a graph architecture (more specifically, EdgeConv~\cite{EdgeConv}) to generate refined feature vectors, where each prototype feature vector is considered as a node. Through this process, contextual information of both near and distant pixels can be effectively leveraged. By plugging the proposed AGCM into the model, structures of a scene can be better appreciated, leading to a better SOD performance.

We validate our proposed AGCM on public benchmark datasets, i.e., DUTS~\cite{wang2017learning}, ECSSD~\cite{yan2013hierarchical}, and DUT-OMRON~\cite{yang2013saliency}. Quantitative and qualitative comparison with other state-of-the-art methods demonstrates the effectiveness of our proposed AGCM. Especially, on complex scenes where multiple objects exist, using AGCM can dramatically boost the performance of a model. A solid ablation study on benchmark datasets confirms the efficacy of AGCM.

Our contributions can be summarized as follows:
\begin{itemize}
\item We propose an AGCM to capture the local structural information and long-range dependencies between distant pixels, enabling the network to leverage global context in an image effectively. 
\item The proposed AGCM can be easily plugged into existing segmentation models without the need for architecture modifications. 
\item Our approach outperforms other state-of-the-art methods on public benchmark datasets quantitatively and qualitatively.
\end{itemize}

\section{Method}
\subsection{Network Overview}
Figure~\ref{figure1} shows an overview of our network architecture. Our model largely consists of an encoder, AGCMs, and a decoder. The input image is fed into the encoder and multi-scale feature maps are extracted. Each feature map is passed through a convolutional layer and connected to the decoder via skip connections. The proposed AGCM is added only for the two highest-level skip connections, taking into account computational cost and memory consumption. To incorporate multi-scale properties, an atrous spatial pyramid pooling~\cite{chen2017deeplab} module is located behind the higher-level AGCM. The decoder progressively upscales the feature maps and generates the final binary prediction mask.

\subsection{Adaptive Graph Convolution Module}
Figure~\ref{figure2}. presents an overview of our proposed AGCM. Using an adaptive graph convolution architecture, AGCM can capture the global context of an image effectively. The AGCM process can be divided into two steps: 1) prototype feature generation; 2) prototype feature refinement. The first stage generates prototype feature vectors from input features, while the second stage refines those feature vectors using graph convolutions. The objective of AGCM is to construct prototype features that contain local structural information and calculate semantic relations between grouped regions, enabling a solid understanding of a given image. 

\begin{figure*}[t]
    \centering
    \includegraphics[width=0.85\linewidth]{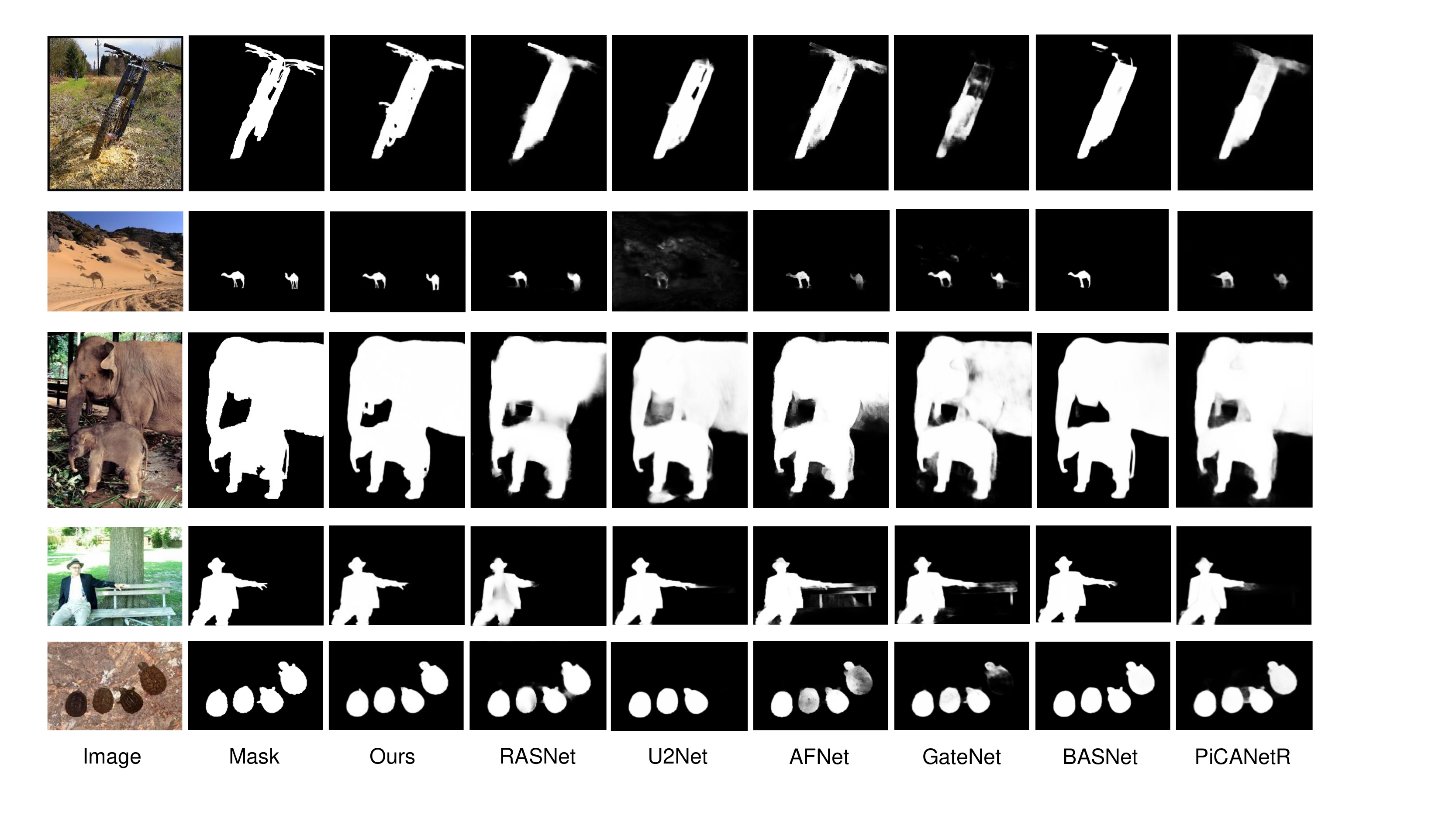}
    \caption{Qualitative comparison of our proposed method to other state-of-the-art methods.}
    \label{figure3}
\end{figure*}

\vspace{1mm}
\noindent\textbf{Prototype feature generation.} 
In the prototype feature generation step, the prototype features $P\in \mathbb{R}^{C\times K}$ are generated from input features $I \in \mathbb{R}^{C\times HW}$, where $C$ and $K$ indicate the channel size and number of feature vectors, respectively. Specifically, attention features $S \in [0,1]^{K\times HW}$ are first generated from input features by applying a convolutional layer and a softmax function along the spatial dimension. Considering the sum of pixel values in each map of $S$ is 1, matrix multiplication of $I$ and $S$ is equivalent to a weighted average pooling operation. In conclusion, the prototype features $P$ can be obtained as
\begin{equation}
P = I \times S^T~.
\end{equation}
As this process is learned end-to-end, $S$  automatically learns to provide adequate structural knowledge for prototype generation. As a result, input features are grouped into semantic regions and global features from those regions are used as a prototype feature that represents the global properties of each region.

\vspace{1mm}
\noindent
\textbf{Prototype feature refinement.} In the second stage, the prototype features $P$ are refined using a self-correlation mechanism and self-attention mechanisms. Drawing inspiration from SPSN~\cite{spsn}, the prototype features are first passed through $N$ EdgeConv~\cite{EdgeConv} layers to produce feature embeddings for each prototype. First, a set of prototype feature vectors, which are closely located in the feature space are connected using K-nearest neighbors algorithm. Then, each node receives semantic cues from other nodes connected to a source node. After the prototype features $P$ is passed through the EdgeConv layers, an adjacency matrix $A \in \mathbb{R}^{K \times K}$ is generated. The values of $A$ indicates the correspondence scores between different prototype features, i.e., graph nodes. Given that $\varphi$ indicates the EdgeConv layers, $A$ can be obtained as
\begin{equation}
A = \varphi(P)^T \times \varphi(P)~.
\end{equation}
Then, based on the affinity scores between different nodes, information from one node is transferred to other nodes. To this end, a kernel weight  $W \in \mathbb{R}^{C}$ is first generated to re-weight each channel of the prototype features $P$. The kernel weight $W$ and re-weighted prototype features $P'\in \mathbb{R}^{C\times K}$ can be calculated as
\begin{equation}
W = MLP(MHA(P))~,
\end{equation}
\begin{equation}
P'_i = W \odot P_i~,
\end{equation}
where MHA and MLP are a multi-head attention layer and a multi-layer perceptron layer. $i$ and $\odot$ indicate a single graph node and Hadamard product, respectively. Finally, the refined features $P''\in \mathbb{R}^{C\times K}$ are generated using the re-weighted prototype features $P'$ and the affinity matrix $A$ as
\begin{equation}
P'' = P' \times Softmax(A)~,
\end{equation}
where $Softmax$ indicates a softmax operation to force the sum of allocation weights of each node to be 1. The refined features $P''$ are then compared to input features, and the correlation scores are used as output values of the AGCM. In our framework, the output correlation scores are concatenated along the channel dimension before the decoding blocks.

\begin{table*}[t]
\centering
\small
\caption{Quantitative comparison of our method with other 6 different SOD models. \textcolor{red}{Red} indicates the best, \textcolor{blue}{blue} indicates the second best performance.}
\vspace{2mm}
\begin{tabular}{c|cccc|cccc|cccc}
\hline
& \multicolumn{4}{c|}{DUTS-TE~\cite{wang2017learning}} &\multicolumn{4}{c|}{ECSSD~\cite{yan2013hierarchical}} &\multicolumn{4}{c}{DUT-OMRON~\cite{yang2013saliency}}\\ \cline{2-13} 
\multirow{-2}{*}{Method} &\multicolumn{1}{c|}{$F_{\beta }$↑} &\multicolumn{1}{c|}{M↓} &\multicolumn{1}{c|}{$E_{\xi}$↑} &$S_{\alpha }$↑  &\multicolumn{1}{c|}{$F_{\beta }$↑} &\multicolumn{1}{c|}{M↓} &\multicolumn{1}{c|}{$E_{\xi}$↑} &$S_{\alpha }$↑ &\multicolumn{1}{c|}{$F_{\beta }$↑} &\multicolumn{1}{c|}{M↓} &\multicolumn{1}{c|}{$E_{\xi}$↑} &$S_{\alpha }$↑\\ 
\hline
RASNet~\cite{RAS} &0.751 &0.059 &0.861 &0.839 &0.889 &0.056 &0.914 &0.893 &0.713 &0.062 &0.846 &0.814\\
U2Net~\cite{qin2020u2} &0.792 &{\color[HTML]{000CFF} 0.045} &0.886 &{\color[HTML]{FE0000} 0.874} &0.892 &{\color[HTML]{FE0000} 0.033} & {\color[HTML]{FE0000} 0.924} &{\color[HTML]{FE0000} 0.928} & {\color[HTML]{FE0000} 0.761}& {\color[HTML]{FE0000} 0.054} & 0.847 & {\color[HTML]{FE0000} 0.871}\\
AFNet~\cite{AFNet} & {\color[HTML]{000CFF} 0.793} & 0.046 & 0.879 & 0.867 & {\color[HTML]{000CFF} 0.908}& 0.042& 0.918 & 0.913& 0.739& 0.057 & 0.853& 0.826\\
GateNet~\cite{Gatenet} & 0.783& {\color[HTML]{000CFF} 0.045} & 0.881& {\color[HTML]{000CFF} 0.870} & 0.896 & \multicolumn{1}{l}{0.041} &{\color[HTML]{000CFF} 0.921}& {\color[HTML]{000CFF} 0.917} & 0.723& 0.061 & 0.848& 0.821\\
BASNet~\cite{Basnet} & 0.791 & 0.048 & {\color[HTML]{000CFF} 0.884}& 0.866 & 0.880 & {\color[HTML]{000CFF} 0.037} & {\color[HTML]{000CFF} 0.921}
& 0.916& 0.756& {\color[HTML]{000CFF} 0.056}&{\color[HTML]{FE0000} 0.869}& {\color[HTML]{000CFF} 0.836}\\
PiCANetR~\cite{picanet} & 0.749 & 0.054 & 0.852 & 0.861& 0.886 & 0.046 & 0.913& {\color[HTML]{000CFF} 0.917} & 0.717& 0.065& 0.841& 0.832 \\
Ours& {\color[HTML]{FE0000} 0.826}& {\color[HTML]{FE0000} 0.039}&{\color[HTML]{FE0000} 0.898}&{\color[HTML]{FE0000} 0.874} & {\color[HTML]{FE0000} 0.914} & {\color[HTML]{000CFF} 0.037} & {\color[HTML]{000CFF} 0.921} & 0.913& {\color[HTML]{000CFF} 0.758}   & {\color[HTML]{FE0000} 0.054} & {\color[HTML]{000CFF} 0.868} & 0.830                        
\\\hline
\end{tabular}
\label{table:tb1}
\end{table*}

\begin{table}[t]
\centering
\caption{Ablation study on the proposed AGCM.}
\vspace{2mm}
\small
\resizebox{\columnwidth}{!}{
\begin{tabular}{cc|cccc|cccc}
\hline
\multicolumn{2}{c|}{Layer \#} & \multicolumn{4}{c|}{ECSSD~\cite{yan2013hierarchical}} & \multicolumn{4}{c}{DUT-OMRON~\cite{yang2013saliency}} \\ 
\hline
4  & 5   & \multicolumn{1}{c|}{$F_{\beta }$↑} & \multicolumn{1}{c|}{M↓} & \multicolumn{1}{c|}{$E_{\xi}$↑} & $S_{\alpha }$↑  & \multicolumn{1}{c|}{$F_{\beta }$↑} & \multicolumn{1}{c|}{M↓} & \multicolumn{1}{c|}{$E_{\xi}$↑} & $S_{\alpha }$↑ \\ 
\hline
   &      & 0.910 & 0.041& 0.918  & 0.870 & 0.748 & \textbf{0.054} & 0.865& 0.821         \\
 \ding{51}  &      & 0.896 & 0.039& 0.920  & 0.910 & 0.746 & \textbf{0.054} & 0.865& 0.822        \\
  \ding{51} &\ding{51}  &\textbf{0.914} &\textbf{0.037}&\textbf{0.921}&\textbf{0.913}&\textbf{0.758}&\textbf{0.054}&\textbf{0.868}&\textbf{0.830} \\ 
\hline
\end{tabular}
}
\label{table:tb2}
\end{table}

\section{Experiment}
\subsection{Experimental Setup}
\textbf{Datasets.} We trained our model using DUTS~\cite{wang2017learning} and conducted the evaluation using three popular datasets: DUTS, ECSSD~\cite{yan2013hierarchical}, and DUT-OMRON~\cite{yang2013saliency}. DUTS is the largest SOD dataset that is divided into DUTS-TR consisting of 10,553 images for training and DUTS-TE consisting of 5,019 images for testing. ECSSD is a dataset comprising 1,000 images, each of which is paired with a corresponding ground truth saliency mask. DUT-OMRON is a dataset comprising 5,168 images. During the training process of our model, we only used the image samples from the DUTS-TR dataset.

\vspace{1mm}
\noindent
\textbf{Evaluation metrics.} We used four metrics to evaluate the SOD performance quantitatively. The F-measure takes into account both precision and recall of the predicted saliency map with respect to the ground truth. MAE is used to evaluate the differences between the ground truths and predictions. E-measure was used to evaluate the similarity between the predicted saliency map and the ground truth. S-measure was used to evaluate the structural similarity between the predicted saliency map and the ground truth. A higher S-measure indicates better structural similarity between the two maps. 

\vspace{1mm}
\noindent
\textbf{Implementation Details.} Our experiments were implemented on two Titan RTX GPUs. We used the Adam~\cite{kingma2014adam} optimizer as optimizer and cosine annealing scheduler to optimize the network. The learning rate is gradually decreased from 1e-4 to 1e-5. The network was trained on the DUTS-TR dataset for 200 epochs with a batch size of 32. As a data augmentation strategy, random horizontal flip was used. Note that all images were resized to a 352$\times$352 pixel resolution for both training and testing. The number of EdgeConv layers, $N$, was set to 3.

\label{sec:expSetup}
\subsection{Comparision with other models}
\noindent\textbf{Qualitative comparison.} In Figure~\ref{figure3}, we qualitatively compare our proposed method to existing state-of-the-art RASNet~\cite{RAS}, U2Net~\cite{qin2020u2}, AFNet~\cite{AFNet}, GateNet~\cite{Gatenet}, BASNet~\cite{Basnet}, and PICANetR~\cite{picanet}. On DUTS-TE~\cite{wang2017learning}. As can be seen from the figure, our new model shows a better mask prediction quality compared to other methods. Especially when images contain multiple objects (e.g., the second, third, and the fifth sequences), it significantly outperforms its competitors. This confirms that the proposed AGCM can effectively detect an image's global context and distinguish objects in the foreground from the background.

\noindent\textbf{Quantitative comparison.} In Table~\ref{table:tb1}, we quantitatively compare our proposed method with existing solutions stated above. On DUTS-TE~\cite{wang2017learning} dataset, the new method outperforms all other methods on all evaluation metrics by a significantly margin. For the ECSSD~\cite{yan2013hierarchical} and DUT-OMRON~\cite{yang2013saliency} datasets, it performed best with the F-measure and the E-measure, respectively. Our new method is based on a simple encoder-decoder architecture and a simple VGG16~\cite{simonyan2014very} encoder.

\subsection{Ablation Study}
We conducted an ablation study on the proposed AGCM and presented the results in Table~\ref{table:tb2}. The study evaluated the performance of the model without any AGCM and with one AGCM module applied to the encoder. We used the ECSSD and DUT-OMRON datasets for the evaluation. The results showed that only one AGCM module plugged into the encoder resulted in slightly increased performance compared to no AGCM, but the overall performance was the best when two AGCM modules were plugged in. The findings demonstrate that incorporating AGCM into the SOD model's encoders, which use the existing VGG encoder, led to significant performance improvements compared to the model without AGCM.

\section{Conclusion}
In this study, we introduced an AGCM that can be easily integrated into existing models. The AGCM combines adaptive graph convolution and edge convolution, which enable the network to better comprehend the image's structural information by aggregating information from both distant and neighboring pixels. Despite its simple architecture, our proposed model obtained better SOD results than existing state-of-the-art methods. In addition, an ablation study backs up the efficacy of our approach to better understand semantic relations between various pixel locations in an image.

\bibliographystyle{IEEEbib}
\bibliography{strings,refs}

\end{document}